\documentclass[sigconf]{acmart}

\usepackage{threeparttable}
\usepackage{bbding}
\usepackage{makecell}
\usepackage{tabularx}
\usepackage{multirow}
\usepackage{balance} 
\usepackage{pdfpages}

\AtBeginDocument{%
  }

\copyrightyear{2025}
\acmYear{2025}
\setcopyright{acmlicensed}
\acmConference[MM '25]{Proceedings of the 33rd ACM International Conference on Multimedia}{}{October 27--31, 2025, Dublin, Ireland}
\acmBooktitle{Proceedings of the 33rd ACM International Conference on Multimedia (MM '25), October 27--31, 2025, Dublin, Ireland}
\acmDOI{10.1145/3746027.3755056}
\acmISBN{979-8-4007-2035-2/2025/10}

\begin{document}

\title{DA-Font: Few-Shot Font Generation via Dual-Attention Hybrid Integration}

\author{Weiran Chen}
\affiliation{%
  \institution{School of Computer Science and Technology, Soochow University}
  \city{Suzhou}
  \country{China}}
\email{wrchen2023@stu.suda.edu.cn}

\author{Guiqian Zhu}
\affiliation{%
  \institution{School of Computer Science and Technology, Soochow University}
  \city{Suzhou}
  \country{China}}
\email{gqzhu@suda.edu.cn}

\author{Ying Li}
\affiliation{%
  \institution{School of Computer Science and Technology, Soochow University}
  \city{Suzhou}
  \country{China}}
\email{ingli@suda.edu.cn}

\author{Yi Ji}
\affiliation{%
  \institution{School of Computer Science and Technology, Soochow University}
  \city{Suzhou}
  \country{China}}
\email{jiyi@suda.edu.cn}

\author{Chunping Liu}
\authornote{Corresponding author}
\affiliation{%
  \institution{School of Computer Science and Technology, Soochow University}
  \city{Suzhou}
  \country{China}}
\email{cpliu@suda.edu.cn}

\renewcommand{\shortauthors}{Weiran Chen, Guiqian Zhu, Ying Li, Yi Ji, \& Chunping Liu}


\begin{abstract}
Few-shot font generation aims to create new fonts with a limited number of glyph references. It can be used to significantly reduce the labor cost of manual font design. However, due to the variety and complexity of font styles, the results generated by existing methods often suffer from visible defects, such as stroke errors, artifacts and blurriness. To address these issues, we propose DA-Font, a novel framework which integrates a Dual-Attention Hybrid Module (DAHM). Specifically, we introduce two synergistic attention blocks: the component attention block that leverages component information from content images to guide the style transfer process, and the relation attention block that further refines spatial relationships through interacting the content feature with both original and stylized component-wise representations. These two blocks collaborate to preserve accurate character shapes and stylistic textures. Moreover, we also design a corner consistency loss and an elastic mesh feature loss to better improve geometric alignment. Extensive experiments show that our DA-Font outperforms the state-of-the-art methods across diverse font styles and characters, demonstrating its effectiveness in enhancing structural integrity and local fidelity. The source code can be found at \href{https://github.com/wrchen2001/DA-Font}{\textit{https://github.com/wrchen2001/DA-Font}}.
\end{abstract}



\maketitle

\section{Introduction}
The task of few-shot font generation allows to transfer the font style from a source domain to a target domain using only a few reference images. It can greatly alleviate the burden of extensive and time-consuming manual design, especially for some character-rich scripts, such as Chinese, Japanese, or Korean. Therefore, few-shot font generation techniques can benefit many critical applications, including logo design, ancient character restoration, and so on.

At present, with the rapid development of deep learning techniques, such as Convolutional Neural Network (CNN)~\cite{1}, Generative Adversarial Network (GAN)~\cite{2} and diffusion model~\cite{3}, researchers have made great advances in creating gratifying fonts. Inspired by neural style transfer~\cite{4}, initial font generation methods mainly focus on learning a mapping function between the source font and the target font. Nevertheless, these approaches can only convert glyphs from one known domain to another that has appeared during training. Hence, to make models extensible to new styles, one popular strategy in recent studies is to disentangle the content and style representations from the given content and reference images. These two representations are then combined and decoded to output the target glyph~\cite{6,8}. In addition, some works also adopt structure-aware representations, which typically decompose characters into different components and acquire multiple style representations~\cite{13, 14} to improve the performance.

Despite the remarkable progress, existing methods still have several drawbacks. The highly diverse and intricate nature of font styles often leads to obvious defects in synthesized results, like incomplete or unwanted strokes, anomalous blurriness and artifacts. Besides, while many approaches are concerned with the use
of component information for glyph decomposition, they usually
ignore the crucial role that components play in feature interactions
between the content and style representations. Considering that
many logographic scripts are inherently component-based, this
neglect would make it difficult to maintain precise glyph shapes,
particularly for some complex characters. 

To tackle the aforementioned challenges, in this paper, we put forward a novel end-to-end few-shot font generation method called DA-Font. To be specific, we design an innovative Dual-Attention Hybrid Module (DAHM). This module is composed of two attention blocks: the component attention block and the relation attention block. The component attention block takes the component-wise codebook of the content image along with the extracted style feature as inputs, and applies a series of operations to produce a stylized component-wise codebook. This representation is then fed into the relation attention block, which further fuses it with the extracted content feature and the raw component-wise codebook. Through this paradigm, the final-obtained representation can effectively harmonize the structural features of the content image with the local attributes of the style pattern. Apart from these, to strengthen the geometric alignment, we bring in two specialized loss functions: corner consistency loss and elastic mesh feature loss. The corner consistency loss ensures the accurate preservation of critical junction points, while the elastic mesh feature loss promotes the topological coherence of the generated characters. Consequently, our method yields superior and impressive stylization performance.

To sum up, the major contributions of this paper are as follows:
\begin{itemize}
    \item We devise a Dual-Attention Hybrid Module (DAHM), which consists of a component attention block and a relation attention block to efficiently integrate content structural features with local stylistic attributes. Based on this, we introduce DA-Font, a new few-shot font generation framework that facilitates style transfer while preserving glyph integrity.
    \item We design two auxiliary loss functions, corner consistency loss and elastic mesh feature loss. These two loss functions optimize discrete junction points and topological coherence respectively, thereby reducing glyph distortions and improving the overall visual consistency.
    \item Extensive experimental results verify that our DA-Font surpasses the leading models in both qualitative and quantitative evaluations, which demonstrates its effectiveness and generalizability on diverse font styles and characters.  
\end{itemize}

\section{Related Work}
\subsection{Image-to-Image Translation}
Image-to-image (I2I) translation is the task of converting a source image to the target domain while maintaining its semantic content. Previous methods mainly utilize GAN~\cite{2} and yield vivid results. Pix2pix~\cite{10} is the first generation model for the I2I translation, which is built upon conditional GAN (cGAN) and leverages paired data for training. However, paired data are often unavailable in many scenarios. Therefore, several methods are proposed to solve this problem. CycleGAN~\cite{16}, DiscoGAN~\cite{17} and DualGAN~\cite{18} introduce the cycle consistency loss to achieve unsupervised I2I translation. Later, inspired by the human ability of inductive reasoning, FUNIT~\cite{4} uses Adaptive Instance Normalization (AdaIN)~\cite{23} to fuse the encoded content and style features. In recent years, diffusion models, a powerful generative technology, have gained wide research attention. ILVR~\cite{24} achieves high-quality performance based solely on a pre-trained unconditional Denoising Diffusion Probabilistic Model (DDPM)~\cite{3}. SCDM~\cite{26} applies stochastic perturbations to semantic maps and conditions I2I translation on the diffused labels. Intuitively, since font generation belongs to a typical I2I translation task, many generic I2I translation models can be adaptively modified for font generation.

\subsection{Few-Shot Font Generation}
Few-shot font generation intends to generate an entire font library in the required style with just a handful of reference glyphs. Early approaches~\cite{5,9,11,15} usually treat this task as an image transformation problem and train cross-domain networks to learn source-to-target mappings. Whereas, these models struggle to generate unseen fonts. To solve this problem, EMD~\cite{6} and AGIS-Net~\cite{8} disentangle the representations of style and content, and model each font as a universal representation. DG-Font~\cite{12} introduces a feature deformation skip connection to capture the glyph deformations. CF-Font~\cite{28} extends DG-Font through incorporating a content fusion module to narrow the gap between the source and target fonts. NTF~\cite{29} formulates font generation as a continuous transformation via a neural transformation field~\cite{30}. FontDiffuser~\cite{32} implements a multi-scale content aggregation block along with a style contrastive refinement module to guide the whole framework. 

\begin{figure*}[h]
  \centering
  \includegraphics[width=\linewidth]{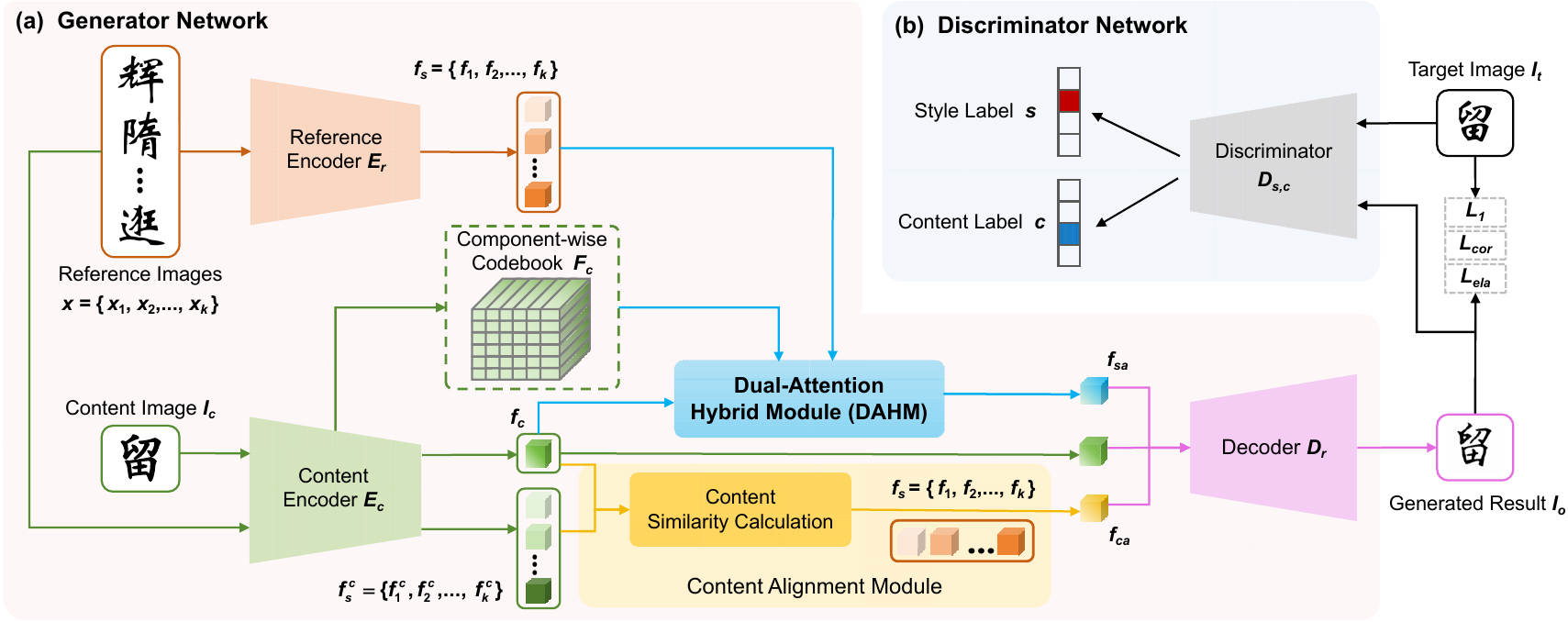}
  \caption{Overview of our proposed DA-Font. (a) The generator network mainly consists of five parts: a pre-trained content encoder $E_c$, a reference encoder $E_r$, a content alignment module, a Dual-Attention Hybrid Module (DAHM) and a decoder $D_r$. (b) A discriminator network is employed to distinguish the real and fake images, while also classifying the content and style categories of the generated characters.}
\label{Main-Model}
\end{figure*}

For the generation of highly-structured characters, some notable researches utilize prior domain knowledge, such as stroke decomposition and trajectory, to optimize the final results. SA-VAE~\cite{35} stands out as the first attempt to integrate radicals and spatial structures into the generative model. LF-Font~\cite{13} represents component-wise style via low-rank matrix factorization. MX-Font~\cite{37} leverages a multi-head design, with each head dedicated to extract distinct local concepts in a weakly supervised manner. CG-GAN~\cite{40} trains a component predictor to better supervise the generator during adversarial training, while FsFont~\cite{41} builds a character-reference mapping relationship and employs cross-attention to align the patch-level features. Diff-Font~\cite{14} infuses predefined embedding tokens into the condition diffusion model to support the sampling process. DP-Font~\cite{7} adopts a multi-attribute guidance and a strict stroke order to direct the generation process. IF-Font~\cite{42} replaces the source image with the Ideographic Description Sequence (IDS) to control the semantics of generated glyphs. Nevertheless, these methods still demand the labels of component categories. In contrast, VQ-Font~\cite{39} constructs a vector quantization-based encoder to automatically extract components. Although existing approaches successfully exploit component information for glyph representation, they often overlook its potential to coordinate the style-content feature interaction, which would easily lead to degraded local fidelity and compromised structural accuracy in the generated glyphs.

\section{The Proposed Method}
In this section, we provide a detailed description of our proposed method. First, we briefly introduce the overall scheme of our method (Sec.~\ref{Overall Scheme}). Next, we present the details of key components in our approach, including glyph feature decomposition and the content alignment module (Sec.~\ref{Content Alignment and Decomposition}), followed by our designed Dual-Attention Hybrid Module (DAHM) (Sec.~\ref{DAHM}). Lastly, we summarize the overall training objective in Sec.~\ref{Training Objective}.

\subsection{Overall Scheme}\label{Overall Scheme}
Given a set of $k$ reference images $x=\{x_1, x_2, ..., x_k\}$ and a content image $I_{c}$, our model aims to generate a character $I_o$ that retains the same content with $I_{c}$ and the same style with $x$. As illustrated in Figure~\ref{Main-Model} (a), the generator network mainly includes five parts. Among them, the reference encoder $E_r$ is used to learn the style representations from the reference images $x$ and map them into a style latent vector $f_s=\{f_1, f_2, ..., f_k\}$. The content encoder $E_c$ processes the content image $I_c$ to extract its feature structure representation $f_c$. It is pre-trained via a Vector Quantized Variational Auto-Encoder (VQ-VAE) to acquire a component-wise codebook $F_c$ as well. Next, our proposed DAHM fuses the component-wise codebook $F_c$, the style latent vector $f_s$, and the content structure feature $f_c$ to derive the style feature $f_{sa}$. Besides, the content alignment module re-weights and aggregates the style feature representation $f_s$ channel-wise to get the aligned content feature $f_{ca}$, with the weight determined by the normalized distance between the content representations of each reference character and the input character. Ultimately, the decoder $D_r$ generates the target image $I_o$ by concatenating the structure feature $f_c$, the style feature $f_{sa}$ and the aligned content feature $f_{ca}$ as inputs.
\begin{figure}[t]
  \centering
  \includegraphics[width=\linewidth]{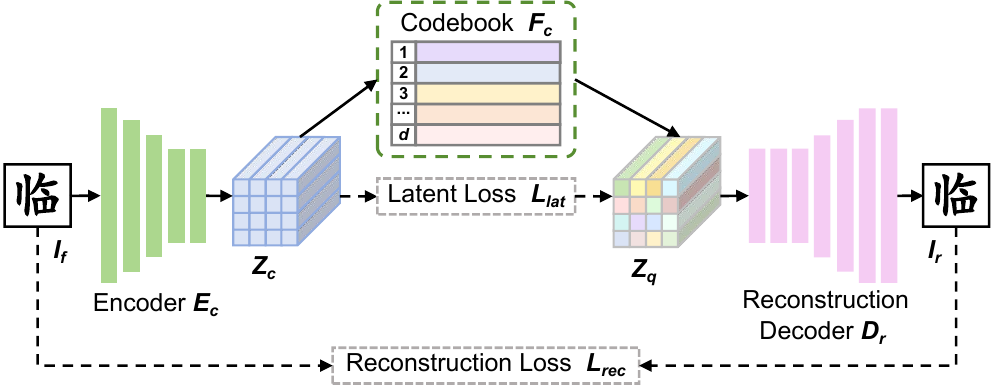}
  \caption{The architecture of glyph feature decomposing network. It is used for pre-training the content encoder $E_c$ and acquiring the component-wise codebook $F_c$.}
\label{VAE}
\end{figure} 

During model's training, a multi-task discriminator $D_{s,c}$ is employed to distinguish between the real and generated images, as shown in Figure~\ref{Main-Model} (b). Moreover, to ensure that the generated glyph accurately reflects the reference style while preserving the input content structure, it also performs the classification on each character, identifying both its style and content categories.
\begin{figure*}[h]
  \centering
  \includegraphics[width=\linewidth]{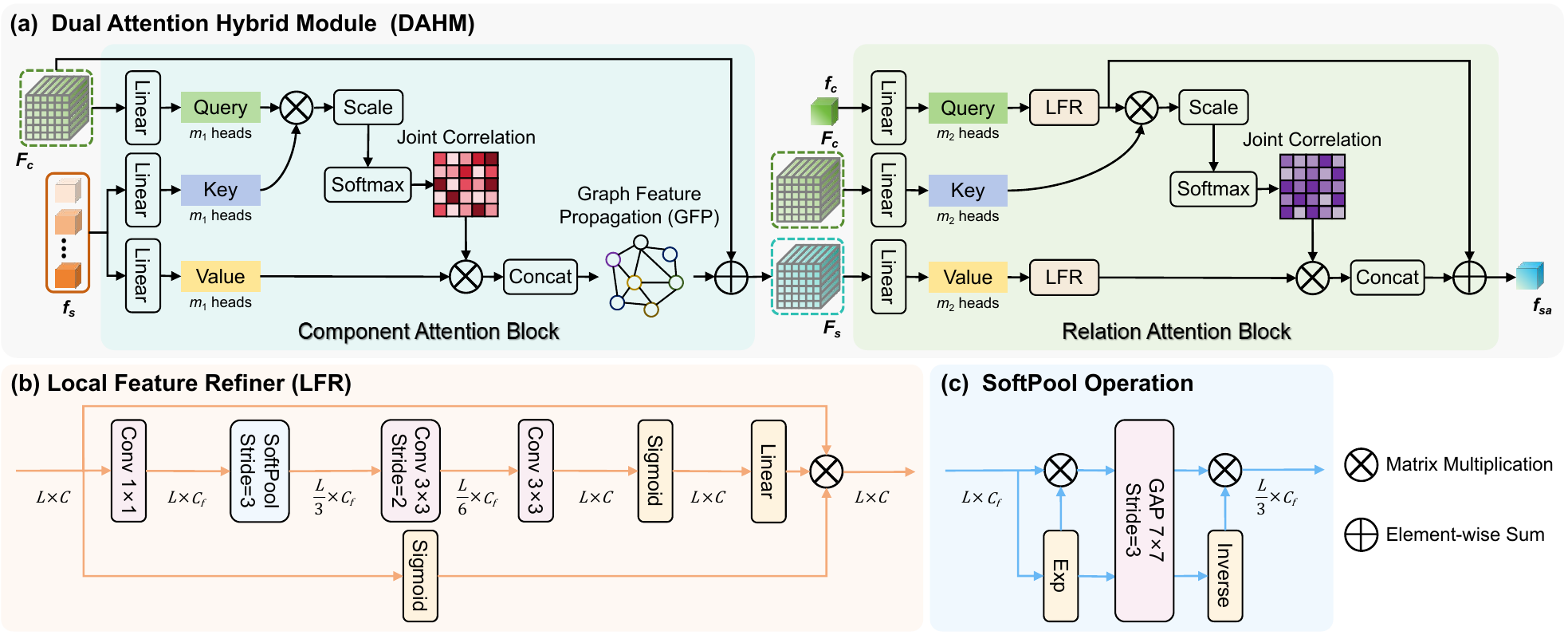}
  \caption{Illustration of Dual Attention Hybrid Module (DAHM). (a) The whole module is comprised of two blocks: component attention block and relation attention block. (b) The Local Feature Refiner (LFR) in the relation attention block optimizes the representation through SoftPool and convolution operations. (c) The SoftPool operation in LFR adjusts the feature map via exponential scaling and pooling.}
\label{Dual-Attention}
\end{figure*}

\subsection{Glyph Decomposition and Alignment}\label{Content Alignment and Decomposition}
\textbf{Glyph Feature Decomposition.} In our method, the content encoder is pre-trained through a glyph feature decomposing network, which decomposes each character into a component-wise codebook. This network is trained on a certain character set for image reconstruction. As illustrated in Figure~\ref{VAE}, the content encoder $E_c$ is built on CNN and maps a character image $I_f$ into latent representation $Z_c$. Then, vector quantization~\cite{47,48} is applied to discrete $Z_c$ by:
\begin{equation}
z_c^i=e_c^j,\quad\mathrm{s.t.}\quad j=\underset{j\in\{1,2,...,d\}}{\operatorname*{\mathrm{arg}\operatorname*{min}}}\|z_c^i-e_c^j\|_2^2
\end{equation}
\noindent{where each spatially-elemental vector $z_c^i$ in $Z_c$ is replaced with the nearest code vector $e_c^j$ from the codebook $F_c$, containing $d$ code vectors. Finally, the reconstruction decoder $D_r$ uses the matched codes $Z_q$ as input and outputs a reconstructed glyph $I_r$.}

During pre-training, $E_c$ and $D_r$ are optimized by minimizing an objective function $\mathcal{L}_{pre}$ that contains a reconstruction loss $\mathcal{L}_{rec}$ and a latent loss $\mathcal{L}_{lat}$, which can be expressed as:
\begin{equation}\begin{aligned}
\mathcal{L}_{pre} & =\mathcal{L}_{rec}+\mathcal{L}_{lat} \\
 & =\|I_f-I_r\|_{1}+\alpha\|\mathrm{sg}[Z_{c}]-Z_{q}\|_{2}^{2}+\beta\|Z_{c}-\mathrm{sg}[Z_{q}]\|_{2}^{2},
\end{aligned}\end{equation}
\noindent{where sg denotes the stop-gradient operator. $\alpha$ and $\beta$ are the two balancing hyper-parameters. Experimentally, we set them as 1 and 0.25, respectively.}

Upon completing pre-training, we fix the content encoder $E_c$ along with the codebook $F_c$ to build the font generation model. Notably, although the reconstruction decoder $D_r$ shares the same architecture with the font generation model’s decoder, the latter is re-trained from scratch during the main training phase.

\textbf{Content Alignment Module.} From a perceptual perspective, reference characters sharing similar elements with the input glyph should be more concerned during the style transfer process~\cite{45}. Therefore, in the content alignment module, we first extract the content features $f_s^c=\{f_1^c,f_2^c,...,f_k^c\}$ from reference images via the content encoder $E_c$. Besides, to ensure dimension compatibility, both $f_s^c$ and $f_c$ are reshaped to $\overline{f_s^c}$ and $\overline{f_c}$. Then, the similarity values are calculated via the normalized cross-correlation measurement~\cite{46} as follows: 
\begin{equation}
\Phi_{ia}=\frac{\langle \overline{f_{i}^{ca}},\overline{f_{c}^{a}} \rangle}{\left\|\overline{f_{i}^{ca}}\right\|*\left\|\overline{f_{c}^{a}}\right\|}, a\in\{1,2,\ldots,N\}
\end{equation}
\noindent{where $a$ refers to the position within the $N$-dimensional channel and $\Phi_{ia}$ is a scalar representing the similarity between the $a$-th channel of the $i$-th reference image and the content image. Next, we normalize these values channel-wise using the softmax function, and apply them to weight the reference style feature representations to form the aligned content feature $f_{ca}$ by:} 
\begin{equation}\begin{gathered}
\overline{\Phi}_{ia}=\mathrm{softmax}({\Phi}_{ia}) \\
f_{ca}=\mathrm{concat}_a\left(\sum_{i=1}^{k}\overline{\Phi}_{ia}f_i^a\right)
\end{gathered}
\end{equation}

\subsection{Dual-Attention Hybrid Module}\label{DAHM}
The main idea behind our proposed Dual-Attention Hybrid Module (DAHM) is to make full use of the component information to improve font generation quality. As depicted in Figure~\ref{Dual-Attention} (a), it is made up of two attention blocks: component attention block and relation attention block.

\textbf{Component Attention Block.} The component attention block aims to generate a stylized component codebook. It is built upon multi-head transformer~\cite{49} with $m_1$ parallel attention heads, where the component codebook $F_c$ serves as the query and the reference style vector $f_s$ provides both the key and value. Specifically, the reference maps $f_s=\{f_{i}\}_{i=1}^{k}$ are reshaped and concatenated along the spatial dimension to form a reference sequence $\tilde{f}_s \in \mathbb{R}^{(k \cdot h \cdot w) \times n}$, where ($h$,$w$) is the resolution of the feature maps, $n$ denotes the number of channels. For the $m$-th attention head, we apply two linear projections $W_{k}^{m}, W_{v}^{m} \in \mathbb{R}^{n\times n_m}$ to transform $\tilde{f}_s$ into the key matrix $K_c^{m}$ and the value matrix $V_c^{m}$, respectively. Meanwhile, the query matrix $Q_{c}^{m}$ is acquired by projecting the codebook $F_c \in \mathbb{R}^{d\times n}$ with a learnable weight $W_{q}^{m} \in \mathbb{R}^{n\times n_m}$ as follows:
\begin{equation}\begin{gathered}
Q_{c}^{m}=F_cW_{q}^{m},\quad Q_{c}^{m} \in \mathbb{R}^{d \times n^m}, \\
K_{c}^{m} = \tilde{f}_s W_{k}^{m},\quad K_{c}^{m}\in\mathbb{R}^{(k\cdot h\cdot w) \times n^{m}}, \\
V_{c}^{m} = \tilde{f}_s W_{v}^{m},\quad V_{c}^{m}\in\mathbb{R}^{(k\cdot h\cdot w) \times n^{m}}.
\end{gathered}\end{equation}

\noindent{where $n^{m}$ represents the hidden dimension of $Q_{c}^{m}, K_{c}^{m}$ and $V_{c}^{m}$. Then, the attention output $A_{c}^{m}$ can be calculated as:}
\begin{equation}
A_{c}^{m} = \mathrm{softmax}\left( \frac{Q_{c}^{m}{K_{c}^{m\top}}}{\sqrt{n^m}}\right)V_{c}^{m}
\end{equation}

After obtaining the output from each attention head, we concatenate them along the channel dimension and employ a linear projection $W_c$ to get the $m_1$-head attention result as follows:
\begin{equation}
F_o=\mathrm{concat}(A_{c}^{1}, A_{c}^{2},..., A_{c}^{m_1})W_c
\end{equation}

To further encapsulate the implicit relationships in $F_o$, we introduce Graph Feature Propagation (GFP). GFP leverages message passing~\cite{50,51} to enhance the component interactions. First, we compute an adjacency matrix $\Omega\in \mathbb{R}^{d\times d}$ to capture the pairwise similarities. For the $i$-th and $j$-th latent component code in $F_o$, their affinity is measured via the dot-product similarity~\cite{36}, followed by softmax normalization:
\begin{equation}
{\Omega}_{ij} = \mathrm{softmax}({e_{o}^i}^\top{e_{o}^j})
\end{equation}

Next, the auxiliary score matrix $S\in \mathbb{R}^{d\times d}$ is derived by concatenating component pairs $[{e_{o}^i}\|{e_{o}^j}]$ and projecting them through a shared weight matrix $W_g$ as follows:  
\begin{equation}
S_{ij} = [{e_{o}^i}\|{e_{o}^j}]W_g
\end{equation}

With the similarity matrix $\Omega$ and score matrix $S$, we can get the refined codebook $\tilde{F}_{o}$ by: 
\begin{equation}
\tilde{F}_{o} = \mathrm{LayerNorm}((\Omega + S)F_o + F_o)
\end{equation}

The final stylized codebook $F_s \in \mathbb{R}^{d\times n}$ is obtained through a residual connection between the original codebook $F_c$ and $\tilde{F}_{o}$:
\begin{equation}
F_s = F_c+\tilde{F}_{o}
\end{equation}

\textbf{Relation Attention Block.} The core of the relation attention block is to adaptively enhance helpful information but weaken useless one through a multi-head attention~\cite{49} with $m_2$ heads. However, unlike the component attention block, where the original codebook $F_c$ serves as the query, here it is used as the key to better capture the relational dependencies between content features and stylized representations. In this design, the content structure representation $f_c$ serves as the query, and the stylized codebook $F_s$ as the value. Formally, we first convert $f_c$ into $\tilde{f}_c\in\mathbb{R}^{(h\cdot w) \times n}$. For the $b$-th attention head, the query matrix $Q_{r}^{b}$, the key matrix $K_{r}^{b}$ and the value matrix $V_{r}^{b}$ are derived by:
\begin{equation}\begin{gathered}
Q_{r}^{b} = \tilde{f}_c \Gamma_{q}^{b},\quad Q_{r}^{b} \in \mathbb{R}^{(h\cdot w) \times n^b}, \\
K_{r}^{b} = F_c \Gamma_{k}^{b},\quad K_{r}^{b}\in\mathbb{R}^{d \times n^{b}}, \\
V_{r}^{b} = F_s \Gamma_{v}^{b},\quad V_{r}^{b}\in\mathbb{R}^{d \times n^{b}}.
\end{gathered}\end{equation}
where $\Gamma_{q}^{b}$, $\Gamma_{k}^{b}$ and $\Gamma_{v}^{b} \in \mathbb{R}^{n \times n^{b}}$ are the learnable linear projections, and $n^{b}$ represents the hidden dimension of $Q_{r}^{b}, K_{r}^{b}$ and $r_{c}^{b}$. 

Inspired by~\cite{52,53}, we introduce Local Feature Refiner (LFR) before computing the attention matrix to refine $Q_{r}^{b}$ and $V_{r}^{b}$, translating them into $\tilde{Q}_{r}^{b}$ and $\tilde{V}_{r}^{b}$, as shown in Figure~\ref{Dual-Attention} (b). To be specific, given an input feature map $X$, the importance value of pixel $x$ within its surrounding region $R$ is computed as:
\begin{equation}\label{I(x)}
I(X)|_x=\sum_{i\in R}\frac{e^{x_i}}{\sum_{j\in R}e^{x_j}}\cdot \Gamma_a
\end{equation}
where $\Gamma_a$ is a learnable weight. As depicted in Figure~\ref{Dual-Attention} (c), we instantiate Eq.~(\ref{I(x)}) via stacking a softpool~\cite{54} followed by $3\times3$ convolution. To further expand the receptive field, we incorporate stride and squeeze convolutions, while sigmoid activation and linear interpolation handle rescaling.

Moreover, to mitigate artifacts brought by stride convolution and linear interpolation, we optimize $I(X)$ with a gating mechanism. Instead of applying extra networks, we choose the first channel map $X_{[0]}$ from the input as a gate, ensuring both simplicity and effectiveness. Hence, the overall LFR can be formulated as:
\begin{equation}
LFR(X)=\sigma(X_{[0]})\odot\psi(\sigma(I(X)))\odot X
\end{equation}
where $\sigma(\cdot)$, $\psi(\cdot)$ and $\odot$ denote sigmoid activation, linear interpolation, and scale-dot, respectively. 

Then, the attention output $A_{r}^{b}$ and the $m_2$-head attention result $F_r$ can be calculated by:
\begin{equation}\begin{gathered}
A_{r}^{b} = \mathrm{softmax}\left( \frac{\tilde{Q}_{r}^{b}{K_{r}^{b\top}}}{\sqrt{n^b}}\right)\tilde{V}_{r}^{b} \\
F_r =\mathrm{concat}(A_{r}^{1}, A_{r}^{2},..., A_{r}^{m_2})\Gamma_r + \tilde{Q}_{r}
\end{gathered}
\end{equation}
where $\Gamma_r$ is a linear projection. Finally, the style representation $f_{sa}\in\mathbb{R}^{h \times w \times n}$ is obtained by reshaping $F_r$ back into the spatial dimensions. Experimentally, we set the number of attention heads to 8 for both two attention blocks.

\begin{table*}[t]
  \caption{\label{Quantitative comparison} Quantitative comparison results on UFUC and SFUC datasets.}
  \centering
  \begin{threeparttable}
    \begin{tabular}{p{1.2cm}|p{3.5cm}|p{1.2cm}p{1.2cm}p{1.2cm}p{1.2cm}p{1.2cm}p{1.6cm}}
    \hline
    \makecell[c]{Dataset} & \makecell[c]{Method} & \makecell[c]{SSIM$\uparrow$} & \makecell[c]{RMSE$\downarrow$} & \makecell[c]{LPIPS$\downarrow$} & \makecell[c]{FID$\downarrow$} & \makecell[c]{L1$\downarrow$} & \makecell[c]{User study$\uparrow$} \\
    \hline
    \makecell[c]{\multirow{7}{*}{UFUC}} & \makecell[c]{FUNIT~\cite{4}} & \makecell[c]{0.6276} & \makecell[c]{0.3356} & \makecell[c]{0.2692} & \makecell[c]{71.3708} & \makecell[c]{0.1376} & \makecell[c]{4.60\%}\\
     & \makecell[c]{MX-Font~\cite{37}} & \makecell[c]{0.6966} & \makecell[c]{0.3159} & \makecell[c]{0.2359} & \makecell[c]{62.7500} & \makecell[c]{0.1235} & \makecell[c]{9.85\%}\\
     & \makecell[c]{DG-Font~\cite{12}} & \makecell[c]{0.6527} & \makecell[c]{0.3238} & \makecell[c]{0.2058} & \makecell[c]{61.6022} & \makecell[c]{0.1241} & \makecell[c]{9.70\%} \\
     & \makecell[c]{LF-Font~\cite{13}} & \makecell[c]{0.6768} & \makecell[c]{0.3110} & \makecell[c]{0.2516} & \makecell[c]{66.8840} & \makecell[c]{0.1190} & \makecell[c]{10.55\%} \\
     & \makecell[c]{CF-Font~\cite{28}} & \makecell[c]{0.6613} & \makecell[c]{0.3102} & \makecell[c]{0.2014} & \makecell[c]{59.7644} & \makecell[c]{0.1169} & \makecell[c]{10.20\%} \\
     & \makecell[c]{VQ-Font~\cite{39}} & \makecell[c]{0.6776} & \makecell[c]{0.3066} & \makecell[c]{0.2157} & \makecell[c]{58.2632} & \makecell[c]{0.1175} & \makecell[c]{12.75\%}\\
     & \makecell[c]{FontDiffuser~\cite{32}} & \makecell[c]{0.6390} & \makecell[c]{0.3579} & \makecell[c]{0.2816} & \makecell[c]{79.4458} & \makecell[c]{0.1530} & \makecell[c]{8.10\%}\\
     & \makecell[c]{IF-Font~\cite{42}} & \makecell[c]{0.6891} & \makecell[c]{0.3130} & \makecell[c]{0.2173} & \makecell[c]{59.6292} & \makecell[c]{0.1038} & \makecell[c]{14.15\%}\\
     & \makecell[c]{DA-Font (Ours)} & \makecell[c]{\textbf{0.7352}} & \makecell[c]{\textbf{0.2854}} & \makecell[c]{\textbf{0.1699}} & \makecell[c]{\textbf{54.1856}} & \makecell[c]{\textbf{0.0872}} & \makecell[c]{\textbf{20.10\%}}\\
    \hline
     \makecell[c]{\multirow{7}{*}{SFUC}} & \makecell[c]{FUNIT~\cite{4}} & \makecell[c]{0.6056} & \makecell[c]{0.3583} & \makecell[c]{0.2777} & \makecell[c]{68.8726} & \makecell[c]{0.1545} & \makecell[c]{4.15\%}\\
     & \makecell[c]{MX-Font~\cite{37}} & \makecell[c]{0.6733} & \makecell[c]{0.3314} & \makecell[c]{0.2197} & \makecell[c]{61.0583} & \makecell[c]{0.1321} & \makecell[c]{9.95\%}\\
     & \makecell[c]{DG-Font~\cite{12}} & \makecell[c]{0.6478} & \makecell[c]{0.3278} & \makecell[c]{0.2006} & \makecell[c]{58.3525} & \makecell[c]{0.1277} & \makecell[c]{10.25\%}\\
     & \makecell[c]{LF-Font~\cite{13}} & \makecell[c]{0.6140} & \makecell[c]{0.3335} & \makecell[c]{0.2726} & \makecell[c]{69.2239} & \makecell[c]{0.1520} & \makecell[c]{10.65\%}\\
     & \makecell[c]{CF-Font~\cite{28}} & \makecell[c]{0.6569} & \makecell[c]{0.3186} & \makecell[c]{0.1974} & \makecell[c]{57.9868} & \makecell[c]{0.1202} & \makecell[c]{11.05\%}\\
     & \makecell[c]{VQ-Font~\cite{39}} & \makecell[c]{0.6414} & \makecell[c]{0.3304} & \makecell[c]{0.2069} & \makecell[c]{56.7299} & \makecell[c]{0.1305} & \makecell[c]{13.70\%}\\
     & \makecell[c]{FontDiffuser~\cite{32}} & \makecell[c]{0.6317} & \makecell[c]{0.3691} & \makecell[c]{0.2910} & \makecell[c]{73.3052} & \makecell[c]{0.1574} & \makecell[c]{7.80\%}\\
     & \makecell[c]{IF-Font~\cite{42}} & \makecell[c]{0.6652} & \makecell[c]{0.3280} & \makecell[c]{0.2221} & \makecell[c]{62.8246} & \makecell[c]{0.1213} & \makecell[c]{14.60\%}\\
     & \makecell[c]{DA-Font (Ours)} & \makecell[c]{\textbf{0.7287}} & \makecell[c]{\textbf{0.3019}} & \makecell[c]{\textbf{0.1792}} & \makecell[c]{\textbf{49.2237}} & \makecell[c]{\textbf{0.1116}} & \makecell[c]{\textbf{17.85\%}} \\
    \hline
    \end{tabular}%
    \end{threeparttable}
\end{table*}%

\subsection{Training Objective}\label{Training Objective}
The loss functions of our proposed model contain five parts: adversarial loss, matching loss, style contrast loss, corner consistency loss and elastic mesh feature loss.

\textbf{Adversarial Loss.} To ensure plausibility in both style and content, we utilize a multi-head discriminator $D_{s,c}$ conditioned on the style label $s$ and content label $c$. The loss function is implemented based on the hinge GAN loss~\cite{33}:
\begin{equation}\begin{gathered}
\begin{split}
\mathcal{L}_{adv}^{D} = & -\mathbb{E}_{I_t \sim p_{data}} min \left(0, -1 + D_{s, c}\left(I_t\right)\right) \\
& -\mathbb{E}_{I_o \sim p_{G}} min \left(0, -1 - D_{s, c}\left(I_o\right)\right)
\end{split} \\
\mathcal{L}_{adv}^{G}= -\mathbb{E}_{I_o \sim p_{G}} D_{s, c}\left(I_o\right)
\end{gathered}
\end{equation}
\noindent{where $p_{data}$ and $p_{G}$ denote the set of real character images and generated character images, respectively.}

\textbf{Matching Loss.} To mitigate mode collapse and enforce the generated character $I_o$ closely resemble to the ground truth $I_t$ at both pixel and feature levels, we apply an $\mathcal{L}_1$ loss on the image and its features, as defined below:
\begin{equation}\begin{gathered}
\mathcal{L}_{img}=\mathbb{E}_{I_t \sim p_{data}}\left[\|I_t-I_o\|_{1}\right] \\
\mathcal{L}_{feat}=\mathbb{E}_{I_t \sim p_{data}}\left[\sum_{m=1}^{M}\left\|D_{s,c}^{(m)}(I_t)-D_{s,c}^{(m)}(I_o)\right\|_{1}\right]
\end{gathered}
\end{equation}
\noindent{where $M$ is the number of layers in $D_{s,c}$ and $D_{s,c}^{(l)}$(·) represents the intermediate feature in the $l$-th layer of $D_{s,c}$.} 

\textbf{Style Contrast Loss.} Since a character can appear in various styles, learning a unified character-level representation is essential~\cite{25}. Hence, to help the style encoder capture inter-class style differences while ignoring reference content, we introduce a style contrastive loss to pull intra-class styles closer and push inter-class styles apart in the embedding space as follows:
\begin{equation}\begin{aligned}&\mathcal{L}_{cst}=\\&-log\left(\frac{exp\left(\sum_{i=1}^dF_s^{i\top} F_{s+}^i\right)}{exp\left(\sum_{i=1}^dF_s^{i\top}F_{s+}^i\right)+\sum_{d_s^-}exp\left(\sum_{i=1}^dF_s^{i\top}F_{s-}^i\right)}\right)\end{aligned}\end{equation}

For a given font style $s$ and its corresponding stylized codebook $F_s\in \mathbb{R}^{d\times n}$, $F_{s+}$ represents the positive pairs with the same style but different content. Conversely, $F_{s-}$ denotes the negative pairs with the same content but different styles. In addition, $d_{s}^{-}$ refers to the number of negative style samples.

\textbf{Corner Consistency Loss.} To improve the geometric consistency of key structural points (such as turning points and endpoints) between generated characters and their ground truth, we employ the Shi-Tomasi corner detection algorithm~\cite{43} to extract salient corner features and design a corner consistency loss. This loss quantifies the spatial alignment between generated and real glyph images using a bidirectional nearest-neighbour matching strategy, and computes the mean distance as follows:
\begin{equation}\begin{aligned}&
\mathcal{L}_{cor} =\\& \mathbb{E}_{I_t \sim p_{data}} \left[ \frac{1}{|c_t| + |c_o|} \left( \sum_{i} \min_{j} ||c^i_t - c^j_o||_2 + \sum_{j}\min_{i} ||c^j_o - c^i_t||_2 \right) \right]
\end{aligned}
\end{equation}
\noindent{where $c_t$ and $c_o$ denote the corner point sets extracted from ground truth and generated image, respectively.}

\textbf{Elastic Mesh Feature Loss.} To preserve the local structural consistency of generated characters while maintaining the global shape integrity, we propose an elastic mesh feature loss. This loss enhances feature alignment by sampling spatially distributed mesh points across the character image and extracting localized patch features~\cite{44}. Specifically, we construct an adaptive elastic mesh that flexibly captures local structural variations, and compute feature similarity between corresponding mesh points in the generated image and ground truth by:
\begin{equation}
\mathcal{L}_{ela} = \mathbb{E}_{I_t \sim p_{data}} \left[ || \phi(I_t) - \phi(I_o) ||_2^2 \right]
\end{equation}
\noindent{where $\phi$ represents the feature extraction process based on the elastic mesh patches.}

\textbf{Overall Objective Loss.} Finally, we optimize DA-Font by the following full objective function:
\begin{equation}
\begin{aligned}
\min\limits_{G}\max\limits_{D}\mathcal{L}_{adv}^D+\mathcal{L}_{adv}^G+\lambda_1\mathcal{L}_{img}+\lambda_2\mathcal{L}_{feat}\\
+\lambda_3\mathcal{L}_{cst}+\lambda_4\mathcal{L}_{cor}+\lambda_5\mathcal{L}_{ela}
\end{aligned}
\end{equation}

Here, $\lambda_1,\lambda_2, \lambda_3, \lambda_4$ and $\lambda_5$ are the weighting hyper-parameters. In our experiments, we empirically set them to 1, 1, 0.1, 0.5 and 0.5.

\begin{figure*}[t]
  \centering
  \includegraphics[width=\linewidth]{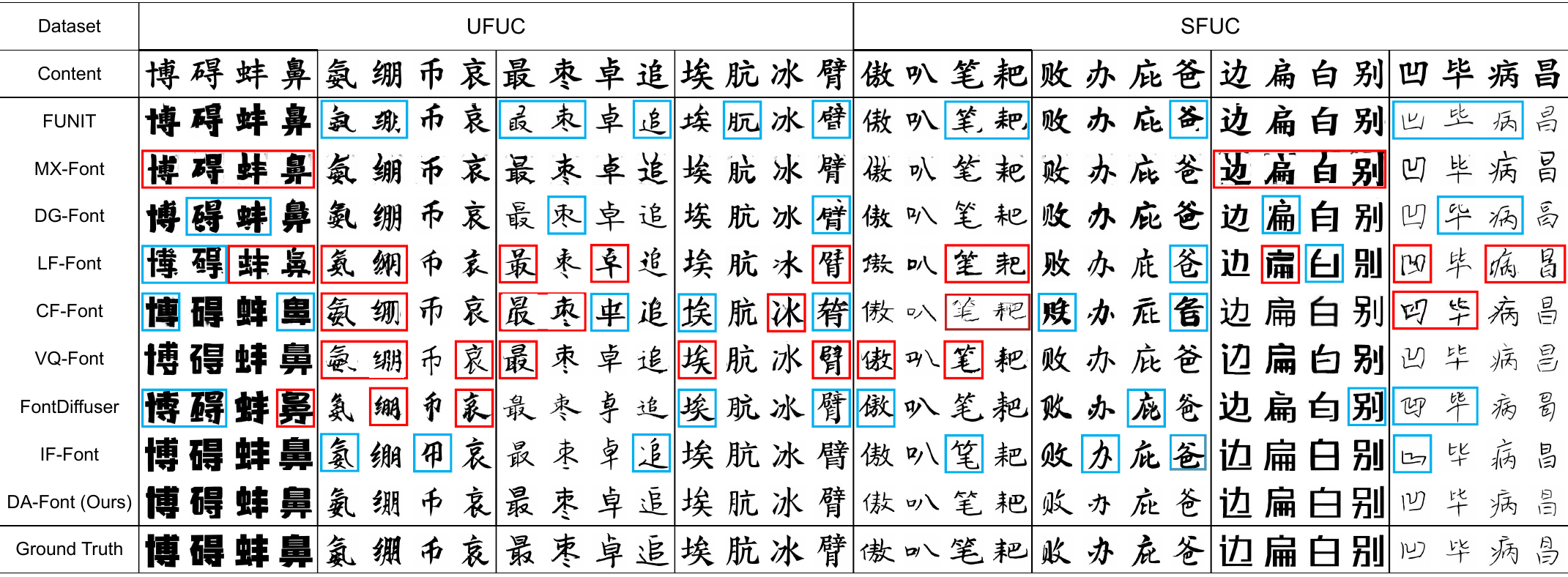}
  \caption{Qualitative comparison results on UFUC and SFUC datasets. Characters in the blue boxes suffer from stroke errors (including missing or redundant strokes), while the red boxes highlight conspicuous blurring or artifacts. Zoom in to see details.}
\label{Qualitative comparison UFUC and SFUC}
\end{figure*} 

\section{Experiments}
\subsection{Dataset and Evaluation Metrics}
\textbf{Dataset.} We collect a Chinese font dataset with 575 fonts (style), each containing 3500 commonly-used Chinese characters (content) at a resolution of 128$\times$128. Notably, the font $kai$ is fixed as the content font throughout training and testing. It is also used to pre-train the feature decomposition network for codebook acquisition.

For the training set, we randomly select 550 fonts with 3000 characters per font, forming the Seen Fonts Seen Characters (SFSC) set. Our test set consists of two parts: 24 Unseen Fonts with 500 Unseen Characters per font (UFUC) and 550 Seen Fonts with 500 Unseen Characters per font (SFUC). 

\textbf{Evaluation Metrics.} To comprehensively assess the quality of font generation, we utilize both pixel-level and perceptual evaluation metrics. Pixel-level metrics, including $L_1$, Root Mean Square Error (RMSE)~\cite{55}, and Structural Similarity Index Measure (SSIM)~\cite{56}, evaluate pixel-wise consistency with ground truth. Perceptual metrics are Learned Perceptual Image Patch Similarity (LPIPS)~\cite{57} and Frechet Inception Distance (FID)~\cite{58}, which measure the feature similarity and are closer to human vision.

Furthermore, we also perform a user study to evaluate the subjective quality of generated results. In particular, we randomly select 10 font styles from each test set and 10 characters per font. Then, we invite 20 well-educated volunteers and ask them to choose the best generation result among the comparison methods. Here, all the samples are shuffled to avoid any potential bias.

\subsection{Implementation Details}
The entire training process consists of two stages. In the first stage, we train the feature decomposing network with 3000 Chinese characters rendered in the font $kai$. During this stage, we set the embedding dimension to 256, the codebook size to 100, the batch size to 64, and the number of training iterations to 50000. In the second stage, we train the entire model via Adam optimizer~\cite{59} with a batch size of 8. Here, the learning rate is set to $2 \times 10^{-4}$ for the generator and $4 \times 10^{-4}$ for the discriminator. The total number of iteration steps is set to 600000. For the few-shot font generation task, the number of reference images is set to 4.

\subsection{Comparison with SOTA Methods}
We compare our model with the following state-of-the-art methods: (1) FUNIT~\cite{4}, a classical few-shot unsupervised image-to-image translation algorithm; (2) MX-Font~\cite{37}, a framework with a multi-head encoder which is trained by weak local component supervision; (3) DG-Font~\cite{12}, a deformable generative network with a feature deformation skip connection module; (4) LF-Font~\cite{13}, a hierarchical framework that analyses localized styles instead of universal styles; (5) CF-Font~\cite{28}, a deep generative model which integrates a content fusion module; (6) VQ-Font~\cite{39}, a hybrid global and local feature style transferring approach; (7) FontDiffuser~\cite{32}, a diffusion-based architecture that models font generation as a noise-to-denoise process; (8) IF-Font~\cite{42}, a paradigm that generates glyphs through ideographic description sequence analysis. For a fair comparison, we retrain all these models using their default settings on our training set.

\begin{table}[t]
    \caption{\label{Loss function} Quantitative results on loss functions.}%
    \centering
    \renewcommand{\arraystretch}{1.16}
    \resizebox{0.45\textwidth}{!}{
\begin{tabular}{cccccc}
\hline
\multicolumn{2}{c}{Loss function} & \multirow{2}{*}{SSIM$\uparrow$} & \multirow{2}{*}{RMSE$\downarrow$} & \multirow{2}{*}{LPIPS$\downarrow$} & \multirow{2}{*}{FID$\downarrow$} \\
$\mathcal{L}_{cor}$          & $\mathcal{L}_{ela}$         &                       &                       &                        &                      \\ \hline
\makecell[c]{\XSolidBrush}        & \makecell[c]{\XSolidBrush}         & 0.7018               & 0.2936                & 0.1945                 & 57.1749              \\  
\makecell[c]{\XSolidBrush}          & \makecell[c]{\CheckmarkBold}       & 0.7247                & 0.2891                & 0.1826                 & 56.9002              \\  
\makecell[c]{\CheckmarkBold}          & \makecell[c]{\XSolidBrush}         & 0.7289                & 0.2873                & 0.1768                 & 55.8974            \\  
\makecell[c]{\CheckmarkBold}          & \makecell[c]{\CheckmarkBold}         & \textbf{0.7352}                & \textbf{0.2854}                &  \textbf{0.1699}                 & \textbf{54.1856}     \\      \hline  
\end{tabular}}
\end{table}

The quantitative comparison results are presented in Table ~\ref{Quantitative comparison}. It is evident that our proposed model achieves the best performance across all the evaluation metrics. This indicates that the generation results of our DA-Font not only have a better fidelity but also align better with human perception. Figure~\ref{Qualitative comparison UFUC and SFUC} displays the qualitative comparison results. From the figure, we can observe that FUNIT exhibits structural incompleteness in most cases. Only when the content font closely resembles the target font can it generate well-layout characters. MX-Font and LF-Font could maintain the overall character shapes, but their outputs often appear blurry with indistinct textures, resulting in a loss of clarity and sharpness. Besides, DG-Font also struggles to capture intricate local details in complex characters, which leads to incomplete components. While CF-Font generally preserves the correct glyph layout, it usually exhibits noticeable artifacts and style inconsistencies. VQ-Font demonstrates good stability but lacks effective control over local features, resulting in glyph distortions and vague characters. The performances of IF-Font and FontDiffuser are also outstanding, but their generated results sometimes suffer from obvious stroke errors. As a whole, characters generated by our DA-Font are of high quality in terms of style consistency, structural correctness and local details. More generation results can be found in the Supplementary Materials.

\subsection{Ablation Study}
In this section, we conduct a series of ablation experiments to access the influence of each part in our model. These experiments are performed on the UFUC dataset. More analyses and results can be found in the Supplementary Materials.

\begin{table}[t]
    \caption{\label{Codebook size} Quantitative results on different codebook sizes.}%
    \renewcommand{\arraystretch}{1.2}
    \centering
    \resizebox{0.45\textwidth}{!}{
    \begin{tabular}{ccccc}
    \hline
        Codebook size & SSIM$\uparrow$ & RMSE$\downarrow$ & LPIPS$\downarrow$ & FID$\downarrow$  \\ \hline
        50 & 0.6219 & 0.3786 & 0.2589 & 69.8627  \\ 
        75 & 0.7064 & 0.3351 & 0.2153 & 63.3152  \\ 
        100 & \textbf{0.7352} & \textbf{0.2854} & \textbf{0.1699} & \textbf{54.1856} \\ 
        125 & 0.7513 & 0.2935 & 0.1608 & 51.1549  \\ 
        150 & 0.7597 & 0.2872 & 0.1686 & 53.6514\\ \hline
    \end{tabular}}
\end{table}

\begin{table}[t]
    \caption{\label{CR} Quantitative results on two attention blocks. C and R denote component attention block and relation attention block, respectively. The first row is the base model.}%
    \renewcommand{\arraystretch}{1.2}
    \centering
    \resizebox{0.45\textwidth}{!}{
    \begin{tabular}{ccccc}
    \hline
       Model & SSIM$\uparrow$ & RMSE$\downarrow$ & LPIPS$\downarrow$ & FID$\downarrow$  \\ \hline
       Base Model & 0.6531 & 0.3197 & 0.2236 & 60.6758 \\ 
        +C & 0.7084 & 0.2982 & 0.1829 & 57.6571  \\ 
        +CR &  \textbf{0.7352} & \textbf{0.2854} & \textbf{0.1699} & \textbf{54.1856}  \\ \hline
    \end{tabular}}
\end{table}

\textbf{The Effect of Different Loss Functions.} To validate the effectiveness of our introduced loss functions, we carry
out ablation experiments on their different combinations. As illustrated in Table~\ref{Loss function}, incorporating either $\mathcal{L}_{cor}$ or $\mathcal{L}_{ela}$ alone could bring in varying degrees of improvement. For instance, with $\mathcal{L}_{ela}$, the relative improvements across SSIM, RMSE, LPIPS, and FID are 0.0229, 0.0045, 0.0119 and 0.2747, respectively. Besides, when both the loss functions are combined, the model achieves the best performance, indicating their complementary roles in optimizing local minutiae and geometric alignments.

\textbf{The Effect of Codebook Size.} The size of codebook $F_c$ plays an important role in determining the model's complexity and feature decomposition ability. Table ~\ref{Codebook size} presents the quantitative results under different codebook sizes. We can see that the performance improves with increasing codebook size up to 100 in both pixel-level and perceptual metrics. However, once the size exceeds 100, the improvements in most metrics plateau, with certain metrics, such as RMSE, even exhibiting a marginal decline. This suggests that while a larger codebook could enhance the representation, excessive sizes would introduce more redundant information, which might compromise the model's efficiency and generalization. Thus, in other experiments, we set the codebook size to 100.

\begin{figure}[t]
  \centering
  \includegraphics[width=\linewidth]{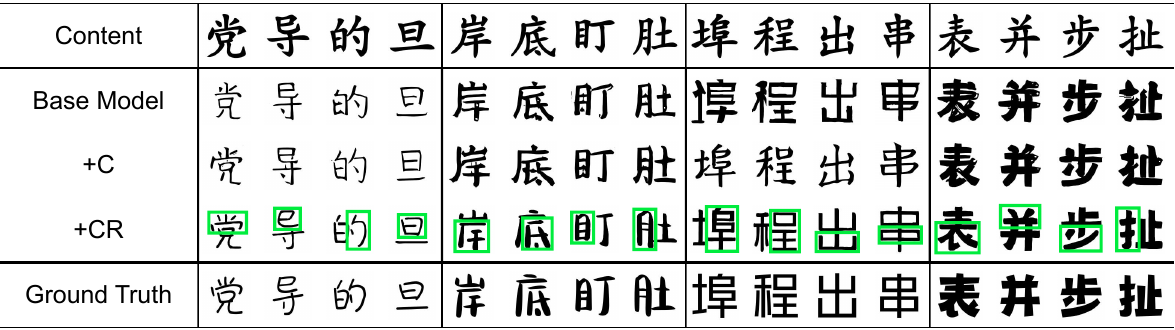}
  \caption{Qualitative results on two attention blocks. C and R are the same notations as Table~\ref{CR}. The green boxes point out details that are better generated by the full model. Zoom in to see details.}
\label{Vision_results_Attention}
\end{figure}

\textbf{The Effect of Each Attention Block.} We separate the two attention blocks in our proposed DAHM and sequentially add them into the base model to observe their individual effects. As depicted in Figure~\ref{Vision_results_Attention}, the base model without both blocks suffers from structural errors and partial blurs. Besides, incorporating the component attention block improves radical alignment and style consistency, yet stroke distortions still exist in some cases. Moreover, the addition of the relation attention bock further strengthens spatial dependencies and local features, leading to better structural coherence. The full model combines these two blocks together, which achieves better performance. The quantitative results in Table~\ref{CR} also confirm the importance of these two blocks.

\begin{figure}[t]
  \centering
  \includegraphics[width=0.8\linewidth]{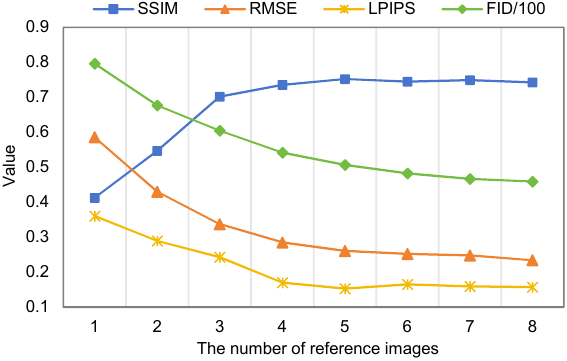}
  \caption{The change trend of model performance with distinct numbers of reference images.}
\label{The effect of reference image numbers.}
\end{figure}

\textbf{The Effect of Reference Image Numbers.} Intuitively, given more reference images, the generated results would be better due to the availability of richer style information. As illustrated in Figure~\ref{The effect of reference image numbers.}, the model’s performance exhibits an upward trend with an increasing number of reference images. Specifically, when the number rises from 1 to 4, the quality of generated results improves significantly. Nevertheless, once the number goes beyond 4, the performance gain becomes marginal. Therefore, in our paper, we set the number of reference images to 4, which could strike a balance between the performance and efficiency. Meanwhile, the model training and comparisons with other models are all completed under this setting.

\section{Conclusion}
In this paper, we propose DA-Font, a novel few-shot font generation framework that incorporates a Dual-Attention Hybrid Module (DAHM). It simultaneously optimizes component-level style transfer and relation-aware feature harmonization, enhancing both the overall structural integrity and local style consistency of the generated characters. Additionally, the integration of corner consistency loss and elastic mesh feature losses further facilitates effective topological fidelity. Both quantitative and qualitative experimental results demonstrate that our proposed model exceeds other competitive methods on various fonts and characters.

In the future, we plan to extend our framework to multilingual font generation, enabling adaptation to diverse script systems beyond Chinese. Furthermore, exploring lightweight deployment for real-time generation may also yield interesting insights.

\begin{acks}
This work was partially supported by Project Funded by the Priority Academic Program Development of Jiangsu Higher Education Institutions, Postgraduate Research \& Practice Innovation Program of Jiangsu Province KYCX24\_3320, and National Natural Science Foundation of China (NSFC Grant No. 62376041).
\end{acks}

\bibliographystyle{ACM-Reference-Format}
\balance
\bibliography{sample-base}

\clearpage


\section*{Supplementary material}

# Supplementary Materials – DA-Font: Few-Shot Font Generation via Dual-Attention Hybrid Integration

**Weiran Chen**
School of Computer Science and Technology, Soochow University
Suzhou, China
wrchen2023@stu.suda.edu.cn

**Guiqian Zhu**
School of Computer Science and Technology, Soochow University
Suzhou, China
gqzhu@suda.edu.cn

**Ying Li**
School of Computer Science and Technology, Soochow University
Suzhou, China
ingli@suda.edu.cn

**Yi Ji**
School of Computer Science and Technology, Soochow University
Suzhou, China
jiyi@suda.edu.cn

**Chunping Liu**[*]
School of Computer Science and Technology, Soochow University
Suzhou, China
cpliu@suda.edu.cn

## 1  More Ablation Studies

Similar to the main paper, we conduct the additional ablation experiments on the UFUC dataset.

**The Effect of GFP and LFR.** To further investigate the effectiveness of our proposed Dual-Attention Hybrid Module (DAHM), we perform several ablation experiments on its two key components: Graph Feature Propagation (GFP) and Local Feature Refiner (LFR). Specifically, we compare the following configurations: (1) both GFP and LFR removed; (2) only GFP or LFR enabled; (3) both GFP and LFR enabled (i.e., the full model).

As shown in Figure 1, when both components are removed, the generated results suffer from structural deformation and style inconsistency. With only GFP enabled, the overall structure becomes more stable, but local regions often appear rough and under-refined. Conversely, enabling LFR alone improves the local appearance and texture clarity, yet the character structure remains distorted and unstable in some cases. Moreover, the full model generates characters that are both structurally coherent and visually sharp, closely resembling the ground truth. The quantitative results in Table 1 also verify the importance of both two components in achieving high-quality font generation under few-shot settings.

**The Effect of $\lambda_4$ and $\lambda_5$ in Loss Functions.** In the overall objective loss, $\lambda_4$ and $\lambda_5$ are two important hyper-parameters associated with our proposed corner consistency loss and elastic mesh feature loss. Hence, we also implement different combinations of $\lambda_4$ and $\lambda_5$, with the other settings held constant. Notably, we use $\lambda_4 + \lambda_5 = 1$ as the control variable for comparison. As shown in Table 2, our model achieves the optimal performance when both $\lambda_4$ and $\lambda_5$ are set to 0.5, indicating that it can provide a better trade-off between the two loss functions and yield more balanced feature learning.

## 2  More Comparison Results

In addition to the two test sets described in the main paper (UFUC and SFUC), we also conduct comparison experiments on a third test set, 24 Unseen Fonts with 3000 Seen Characters per font (UFSC), to further evaluate the generalization ability of our proposed method.

Apart from the state-of-the-art methods mentioned in the main paper, we also include Diff-Font, a generative architecture built

**Figure 1: Qualitative results on two key components in DAHM. The green boxes point out details that are better generated by our full proposed model. Zoom in to see details.**

**Table 1: Quantitative results on two key components.**

| Model | SSIM↑ | RMSE↓ | LPIPS↓ | FID↓ |
|---|---|---|---|---|
| w/o (GFP & LFR) | 0.6892 | 0.3014 | 0.2016 | 57.6539 |
| w/o GFP | 0.7188 | 0.2942 | 0.1815 | 55.1893 |
| w/o LFR | 0.7205 | 0.2923 | 0.1778 | 54.8157 |
| Full Model | **0.7352** | **0.2854** | **0.1699** | **54.1856** |

**Table 2: Quantitative results on different $\lambda_4$ and $\lambda_5$.**

| $\lambda_4$ | $\lambda_5$ | SSIM↑ | RMSE↓ | LPIPS↓ | FID↓ |
|---|---|---|---|---|---|
| 0.2 | 0.8 | 0.6895 | 0.3062 | 0.2014 | 58.2591 |
| 0.3 | 0.7 | 0.6932 | 0.3027 | 0.1894 | 57.2150 |
| 0.4 | 0.6 | 0.7256 | 0.2893 | 0.1751 | 55.1274 |
| 0.5 | 0.5 | **0.7352** | **0.2854** | **0.1699** | **54.1856** |
| 0.6 | 0.4 | 0.7148 | 0.2941 | 0.1827 | 56.3489 |
| 0.7 | 0.3 | 0.7029 | 0.2985 | 0.1968 | 57.9412 |
| 0.2 | 0.9 | 0.6905 | 0.3051 | 0.2003 | 58.1276 |

upon the diffusion model, as an extra comparison method. It is worth noting that since Diff-Font is only capable of generating seen characters, we only evaluate this model on the UFSC test set.

The quantitative comparison results are displayed in Table 3, which shows that our DA-Font consistently outperforms all the methods across multiple evaluation metrics. Figure 2 presents the qualitative comparison results. From the figure, we can see that

**Table 3: Quantitative comparison results on UFSC dataset.**

| Method | SSIM↑ | RMSE↓ | LPIPS↓ | FID↓ | L1↓ | User study↑ |
|---|---|---|---|---|---|---|
| FUNIT | 0.6309 | 0.3339 | 0.2685 | 59.3728 | 0.1362 | 4.05% |
| MX-Font | 0.7062 | 0.3229 | 0.2063 | 53.3935 | 0.1205 | 9.45% |
| DG-Font | 0.6658 | 0.3138 | 0.1877 | 50.0568 | 0.1256 | 9.55% |
| LF-Font | 0.6812 | 0.3090 | 0.2471 | 55.2930 | 0.1176 | 9.85% |
| CF-Font | 0.6794 | 0.3063 | 0.1836 | 49.7423 | 0.1224 | 10.00% |
| VQ-Font | 0.6842 | 0.2995 | 0.1927 | 48.6191 | 0.1166 | 12.65% |
| Diff-Font[1] | 0.6143 | 0.3242 | 0.1851 | 60.0310 | 0.1370 | 7.05% |
| FontDiffuser | 0.6430 | 0.3433 | 0.2846 | 63.1833 | 0.1483 | 6.85% |
| IF-Font | 0.6915 | 0.2997 | 0.1962 | 48.3104 | 0.1181 | 13.60% |
| DA-Font (Ours) | **0.7544** | **0.2782** | **0.1649** | **43.8763** | **0.0916** | **16.95%** |

[1] Diff-Font can only generate seen characters, so we evaluate this model on the UFSC test set alone.

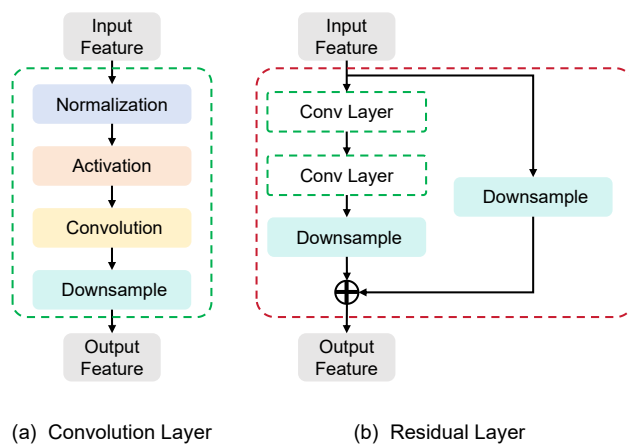

Figure 2: Qualitative comparison results on UFSC dataset. Characters in the blue boxes suffer from stroke errors, while the red boxes highlight conspicuous blurring or artifacts. Zoom in to see details.

FUNIT tends to produce structurally incomplete glyphs, while MX-Font and LF-Font preserve general shapes but often generate blurry outputs with unclear textures. DG-Font struggles with local details, leading to missing strokes or components in complex characters. Besides, CF-Font maintains the overall layout but suffers from artifacts and inconsistent styles. VQ-Font achieves stable style transfer yet lacks fine-grained control, resulting in glyph distortions. Although Diff-Font, FontDiffuser, and IF-Font perform relatively well, stroke-level errors still occur in some cases. In contrast, our DA-Font generates characters with superior style consistency, better structural accuracy, and refined local details.

## 3 Network Architecture

Our entire model is made up of two parts: the generator and the discriminator. Both of them are built up with two types of layers: the Convolution layer and the Residual layer, as illustrated in Figure 3. Notably, the Residual layer contains two identical Convolution layers. Since the Residual layer has its own downsampling operator, its Convolution layers skip the downsampling step. The detailed architectures are shown in Table 4.

Figure 3: Structures of the Convolution Layer and the Residual Layer in Table 4.

(a) Convolution Layer  (b) Residual Layer

## 4 More Generation Results

To further demonstrate the robustness and generalization capability of our model, we conduct two extra groups of experiments. The first group focuses on font generation across diverse and highly stylized typefaces to evaluate the model's adaptability to substantial style variations (Figure 4). The second group targets the generation of structurally complex characters, which contain dense strokes and intricate component arrangements, to verify the model's ability in preserving fine-grained structural fidelity under challenging conditions (Figure 5). Experimental results from both settings consistently validate the effectiveness of our approach in handling a broad spectrum of font generation scenarios.

## 5 Broader Impact

DA-Font enhances the efficiency and creativity of font designers by enabling rapid and high-quality style transfer. In addition, it holds cultural value by supporting the preservation and revitalization of traditional Chinese calligraphy, making classical aesthetics more accessible in digital contexts. Through bridging AI technology with artistic practices, DA-Font promotes the inheritance and innovation of typographic art in the modern era.

**Table 4:** Architectures of the generator modules $E_c$, $E_r$, $D_r$, and the discriminator $D_{s,c}$. IN and SN denote instance normalization and spectral normalization, respectively. All the padding operations are zero-padding. In the discriminator's output layer, we use two embedding operators to embed the output feature map into two prediction vectors of both its style and content.

| **Content Encoder $E_c$** | | | | | | | |
|---|---|---|---|---|---|---|---|
| Layer Type | Normalization | Activation | Padding | Kernel Size | Stride | Downsample | Feature Maps |
| Convolution Layer | IN | ReLU | 1 | 3 | 1 | - | 32 |
| Convolution Layer | IN | ReLU | 1 | 3 | 2 | - | 64 |
| Convolution Layer | IN | ReLU | 1 | 3 | 2 | - | 128 |
| Convolution Layer | IN | ReLU | 1 | 3 | 2 | - | 256 |
| Convolution Layer | IN | ReLU | 1 | 3 | 1 | - | 256 |

| **Reference Encoder $E_r$** | | | | | | | |
|---|---|---|---|---|---|---|---|
| Layer Type | Normalization | Activation | Padding | Kernel Size | Stride | Downsample | Feature Maps |
| Convolution Layer | IN | ReLU | 1 | 3 | 1 | - | 32 |
| Convolution Layer | IN | ReLU | 1 | 3 | 1 | AvgPool | 64 |
| Convolution Layer | IN | ReLU | 1 | 3 | 1 | AvgPool | 128 |
| Residual Layer | IN | ReLU | 1 | 3 | 1 | - | 128 |
| Residual Layer | IN | ReLU | 1 | 3 | 1 | - | 128 |
| Residual Layer | IN | ReLU | 1 | 3 | 1 | AvgPool | 256 |
| Residual Layer | IN | ReLU | 1 | 3 | 1 | - | 256 |
| Output Layer | - | Sigmoid | - | - | - | - | 256 |

| **Decoder $D_r$** | | | | | | | |
|---|---|---|---|---|---|---|---|
| Layer Type | Normalization | Activation | Padding | Kernel Size | Stride | Upsample | Feature Maps |
| Residual Layer | IN | ReLU | 1 | 3 | 1 | - | 256 |
| Residual Layer | IN | ReLU | 1 | 3 | 1 | - | 256 |
| Residual Layer | IN | ReLU | 1 | 3 | 1 | - | 256 |
| Convolution Layer | IN | ReLU | 1 | 3 | 1 | Nearest | 128 |
| Convolution Layer | IN | ReLU | 1 | 3 | 1 | Nearest | 64 |
| Convolution Layer | IN | ReLU | 1 | 3 | 1 | Nearest | 32 |
| Convolution Layer | IN | ReLU | 1 | 3 | 1 | - | 1 |
| Output Layer | - | Sigmoid | - | - | - | - | 1 |

| **Discriminator $D_{s,c}$** | | | | | | | |
|---|---|---|---|---|---|---|---|
| Layer Type | Normalization | Activation | Padding | Kernel Size | Stride | Upsample | Feature Maps |
| Convolution Layer | SN | - | 1 | 3 | 2 | - | 32 |
| Residual Layer | SN | ReLU | 1 | 3 | 1 | AvgPool | 64 |
| Residual Layer | SN | ReLU | 1 | 3 | 1 | AvgPool | 128 |
| Residual Layer | SN | ReLU | 1 | 3 | 1 | AvgPool | 256 |
| Residual Layer | SN | ReLU | 1 | 3 | 1 | - | 512 |
| Residual Layer | SN | ReLU | 1 | 3 | 1 | - | 512 |
| Output Layer | Multi-task embedding | | | | | | |

**Figure 4: More generation results on various font styles. The first row shows the content images, the second row denotes the characters generated by our DA-Font, and the third row represents the ground truth. Zoom in to see details.**

Figure 5: More generation results on complex characters. The first row shows the content images, the second row denotes the characters generated by our DA-Font, and the third row represents the ground truth. Zoom in to see details.



\end{document}